\documentclass[letterpaper]{article} 
\usepackage{aaai24}  
\usepackage{times}  
\usepackage{helvet}  
\usepackage{courier}  
\usepackage[hyphens]{url}  
\usepackage{graphicx} 
\usepackage{natbib}  
\usepackage{caption} 
\usepackage{algorithm}
\usepackage{algorithmic}
\usepackage{xcolor}
\usepackage{comment}
\usepackage{mathtools}
\usepackage{adjustbox}
\usepackage{multirow}
\usepackage{subcaption}
\usepackage{amsmath}
\DeclareMathOperator*{\argmax}{arg\,max}

\usepackage{newfloat}
\usepackage{listings}
\DeclareCaptionStyle{ruled}{labelfont=normalfont,labelsep=colon,strut=off} 
\floatstyle{ruled}
\newfloat{listing}{tb}{lst}{}
\floatname{listing}{Listing}
\title{Towards Machine Unlearning Benchmarks:\\Forgetting the Personal Identities in Facial Recognition Systems}
\author {
    Dasol Choi$^{*1}$ \;
    \stepcounter{footnote}Dongbin Na$^{*2}$\thanks{Correspondence to dongbinna@postech.ac.kr}
}
\affiliations{
    $^1$Kyunghee University \;
    $^1$MODULABS \;
    $^2$Pohang University of Science and Technology (POSTECH)
}

\usepackage{bibentry}

\begin{document}

\maketitle

\def\thefootnote{*}\footnotetext{These authors contributed equally to this work.}

\begin{abstract}

Machine unlearning is a crucial tool for enabling a classification model to forget specific data that are used in the training time.
Recently, various studies have presented machine unlearning algorithms and evaluated their methods on several datasets.
However, most of the current machine unlearning algorithms have been evaluated solely on traditional computer vision datasets such as CIFAR-10, MNIST, and SVHN.
Furthermore, previous studies generally evaluate the unlearning methods in the class-unlearning setup.
Most previous work first trains the classification models and then evaluates the machine unlearning performance of machine unlearning algorithms by forgetting selected image classes (categories) in the experiments.
Unfortunately, these class-unlearning settings might not generalize to real-world scenarios.
In this work, we propose a machine unlearning setting that aims to unlearn specific instance that contains personal privacy (identity) \textbf{while maintaining the original task} of a given model.
Specifically, we propose two machine unlearning benchmark datasets, \textit{MUFAC} and \textit{MUCAC}, that are greatly useful to evaluate the performance and robustness of a machine unlearning algorithm.
In our benchmark datasets, the original model performs facial feature recognition tasks: face age estimation (multi-class classification) and facial attribute classification (binary class classification), where a class does not depend on any single target subject (personal identity), which can be a realistic setting.
Moreover, we also report the performance of the state-of-the-art machine unlearning methods on our proposed benchmark datasets.
All the datasets, source codes, and trained models are publicly available at \textcolor{blue}{\textbf{\url{https://github.com/ndb796/MachineUnlearning}}}.

\end{abstract}

\section{Introduction}

With the explosive growth of the computational power and scale of the datasets, the deep-learning models have shown remarkable achievements in various research fields and industries~\cite{ResNet,EfficientNet,Transformer}.
Furthermore, numerous large-scale foundation models utilizing the scalable architectures~\cite{Transformer} have achieved improved classification performance by training the models on billions of data samples~\cite{BERT,GPT,GPTUnderstand,VITSurvey,LLAMA}.
However, some studies point out that deep neural networks can leak the personal information of some training samples~\cite{MembershipInferenceAttack,ExtractingAttack,DataExtraction2}.
Recent large-scale foundation models are known to be also vulnerable to data extraction attacks due to their heavyweight weight parameters~\cite{ExtractingAttack}.
Moreover, for the \textit{right to be forgotten}, the model providers can be required to remove someone's personal identity that might be used for training an AI-based system in the training time.
Therefore, recent studies have presented various machine unlearning algorithms to efficiently unlearn the samples to be forgotten given a trained model.

However, the most recently proposed milestone studies~\cite{Zeroshot,SISA,NTK,UNSIR,BadTeacher} have considered \textbf{only} the class-unlearning setting with toy computer vision datasets including CIFAR-10~\cite{CIFAR10}, SVHN~\cite{SVHN}, and MNIST.
We note that the class-unlearning setting sometimes could not capture real-world scenarios.
In the class-unlearning setting, they assume that the \textit{target to unlearn} wholly occupies a specific class (category).
This setting does not consider the case of removing a single instance (a specific car image of \textit{car} class) from the original model while maintaining the \textit{car} class.
After the class-unlearning phase, some classes will be entirely removed from the original model $\theta_{original}$.
Therefore, class-unlearning can not be applied to even simple binary classification tasks.
This issue is induced by the circumstances that the task the model solves is directly linked to the target \textit{subject to unlearn} in the class-unlearning setting.

\begin{figure*}[htp]
\hspace*{-0.8cm}
    \centering
\captionsetup[subfigure]{justification=centering}
\subcaptionbox{Comparison with the traditional class-unlearning.}
      [.5\linewidth]
      {\includegraphics[scale=0.5]{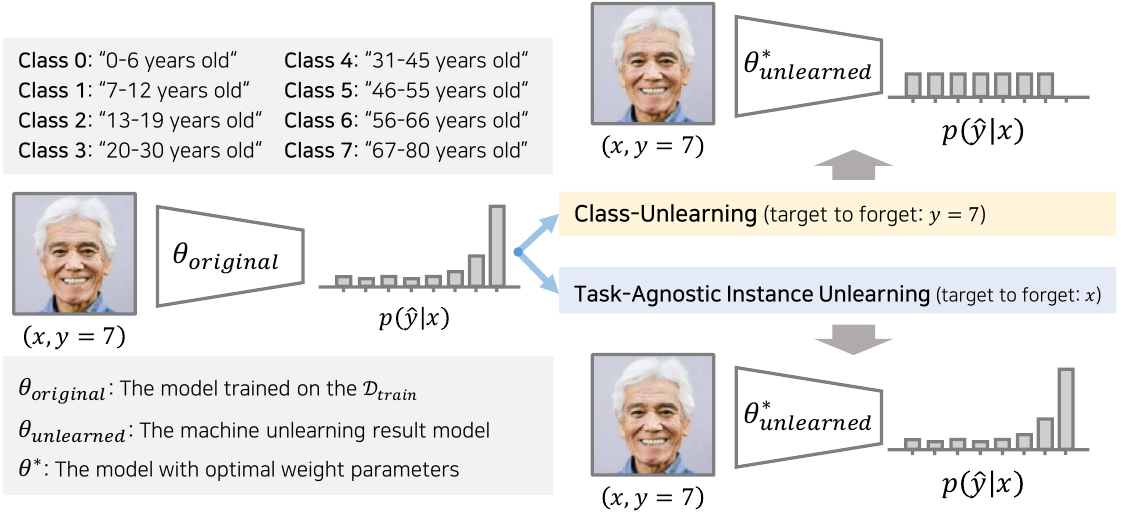}}
\hspace*{\fill}
\subcaptionbox{An illustration of the task-agnostic machine unlearning.}
      [.49\linewidth]
      {\includegraphics[scale=0.53]{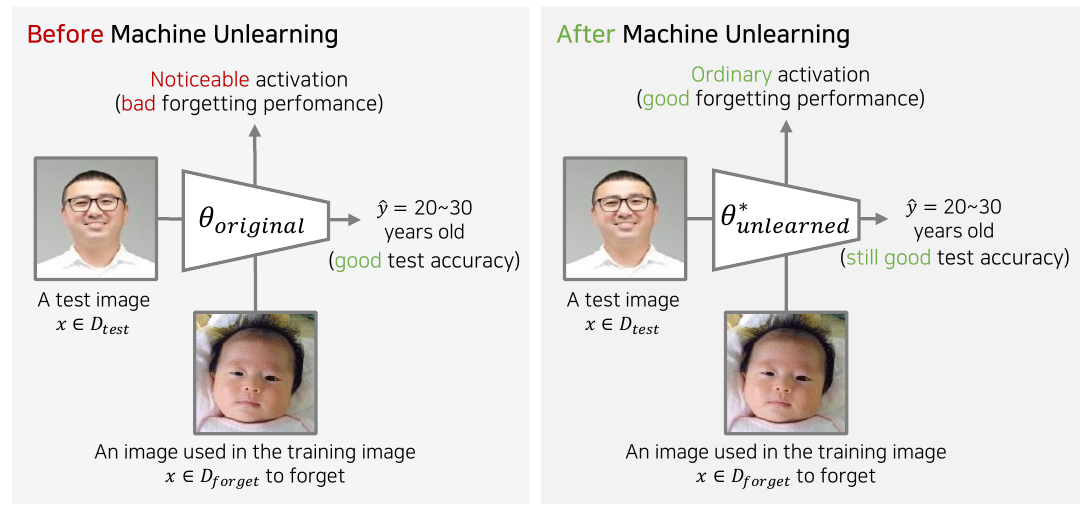}}
    \caption{The illustration of our proposed task-agnostic machine unlearning benchmarks. Figure (a) shows a conceptual illustration of our proposed task-agnostic unlearning setup compared with the class-unlearning setting. Figure (b) represents that our task-agnostic unlearning setup pursues to achieve good accuracy for the original task even after the unlearning is performed.}
    \label{fig:machine_unlearning_concept}
\end{figure*}

In this paper, we present two benchmark datasets for the machine-unlearning research domain.
Both two benchmark datasets address facial recognition tasks.
Firstly, we publicly present a new dataset, \textbf{Machine Unlearning for Facial Age Classifier (MUFAC)}, which is a novel dataset that contains more than 13,000 Asian facial images.
All the face images have been collected from South Korea through the participants.
In this dataset, all the image data samples also provide a corresponding annotation file that includes the labels of (1) age and (2) personal identity number.
The original task of our \textit{MUFAC} is the age classification task, therefore, we train the original classification model $\theta_{original}$ using these face images and the label information of \textit{age}.
In this dataset, the target subject to forget can be a specific person.
We note that \textit{the original task of the unlearned model $\theta_{unlearned}$ will be not changed} after we make the original model $\theta_{original}$ unlearn a subject to be forgotten.
We define this setting as a \textbf{task-agnostic machine unlearning} setting as shown in Figure~\ref{fig:machine_unlearning_concept} (b).
Moreover, we also present a benchmark dataset, \textbf{Machine Unlearning for Celebrity Attribute Classifier (MUCAC)}, that contains 30,000 celebrity facial images.
This dataset is curated from the CelebA dataset~\cite{CelebA} that has already been largely utilized in the face generation research area.
In the MUCAC, all images also represent only the face of a person.
We set the original problem of \textit{MUCAC} to the multi-label classification, predicting the three types of representative facial recognition tasks.
Specifically, we utilize three binary labels, i.e., male/female, old/young, and smiling/unsmiling respectively for each classification task.
All images also contain the personal identity number in \textit{MUCAC}, thus, we also provide the machine unlearning benchmark that aims to make the $\theta_{original}$ forget specific personal identities.
We set the resolution for all images of \textit{MUFAC} and \textit{MUCAC} to 128$\times$128 and the images contain only the part of the face in the human body.
Our contributions are listed as follows:

\vspace{-0.2cm}
\begin{itemize}
    \item We introduce two benchmark datasets, \textit{MUFAC} and \textit{MUCAC} to evaluate the performance of a machine unlearning algorithm. To the best of our knowledge, we are the first to provide the facial age classification dataset \textit{with all the personal identity labels}, which can be especially useful to evaluate the robustness of machine unlearning algorithms.
    \item With extensive experiments, we also report the performance of the state-of-the-art machine unlearning methods that have been previously presented.
    \item In this work, we demonstrate that some state-of-the-art methods that have been reported as showing good machine unlearning performance in the class-unlearning setting might sometimes show poor machine unlearning performance on our proposed new benchmarks that adopt the task-agnostic machine unlearning setup.
    \item We provide all datasets, codes, and trained models publicly for the growth of machine-unlearning research.
\end{itemize}

\section{Related Work}

Although machine unlearning has attracted attention recently, most of the machine unlearning methods still adopt only the toy datasets that address class-unlearning settings.
For example, the recently presented methods including SISA~\cite{SISA}, NTK~\cite{NTK}, and UNSIR~\cite{UNSIR,Zeroshot} all consider class-unlearning benchmark datasets including CIFAR-10~\cite{CIFAR10}, SVHN~\cite{SVHN}, and MNIST.
We claim that unlearning a specific digit from 0 to 9 for MNIST does not make sense in human perception and, thus, might be not realistic in a real-world setting.
In other words, the class-unlearning setting assumes that a specific class is totally related to the images to be unlearned.
If they consider the \textit{personal writing style of digits} of MNIST, the personal identity might be leaked.
However, unfortunately, their work does not consider this case at all.
As a result, in this work, we point out that these toy datasets might not generalize to real-world deployment scenarios.
We provide two task-agnostic machine unlearning benchmark datasets that might capture more realistic machine unlearning scenarios.
Furthermore, we have observed some SOTA methods might not properly perform on our more realistic benchmark setting.
Unlike previous work, our proposed benchmark focuses on facial recognition tasks.
Moreover, the target to forget is the \textit{personal identity}, not the \textit{a specific class}.

\section{Proposed Benchmarks}

We introduce two datasets, \textit{MUFAC} and \textit{MUCAC}, as new machine unlearning benchmark datasets.

\subsection{Task-Agnostic Machine Unlearning}

The aforementioned recent machine unlearning methods consider the case that the samples to unlearn directly indicate a specific class.
Instead of this class-unlearning setting, our work addresses the \textit{task-agnostic machine unlearning} setting.
We assume that a user can request the AI-based service provider to unlearn the specific personal identity that might have been used in the training phase.
In this setting, we consider the machine unlearning process does not change the original task.
Thus, our task-agnostic machine unlearning aims to forget a set of instances, rather than removing some classes totally from the original model $\theta_{original}$.
For example, we can train an age estimation model that predicts the age given facial image input data.
In our setup, forgetting a specific personal identity must not change the original task.
Moreover, the original model utility (classification accuracy) is still good after the unlearning phase, if the forgetting method is effective.
The difference between the traditional class-unlearning setup and our task-agnostic machine-unlearning setup is illustrated in Figure~\ref{fig:machine_unlearning_concept} (a).

\subsection{Evaluation Metrics}

The goal of machine unlearning is generally divided into two aspects: (1) model utility and (2) forgetting performance.

\subsubsection{Model Utility}

First, we should pursue obtaining a good generalization performance for the original task.
In the facial classification system, the classification \textit{accuracy} is considered as the model utility measure.
Accuracy is a measurement that calculates the probability that the predicted value $\hat{y}$ is the ground-truth value $y$ in the test dataset $\mathcal{D}_{test}$.
\vspace{-0.1cm}
\begin{align}
P(\hat{y} = y)
\end{align}

\subsubsection{Forgetting Score}

Machine unlearning aims to unlearn the specific images to forget.
Thus, evaluating the magnitude of the forgetting properly is important.
Previous studies generally utilize accuracy or membership inference attack~\cite{UNSIR} to calculate the degree of forgetting after applying machine unlearning algorithms.
The originally trained model $\theta_{original}$ is basically trained on the $\mathcal{D}_{train}$, thus, the loss values for $x_{train}$ of the $\mathcal{D}_{train}$ are frequently lower than the loss values for an unseen data $x_{unseen}$ of the $\mathcal{D}_{unseen}$.
Therefore, we adopt the MIA (Membership Inference Attack) for evaluating the forgetting performance of a given machine unlearning algorithm.
For this purpose, in the final evaluation phase, we train an additional binary classification model $\psi(\cdot)$ to distinguish the loss value of the $x$ forgotten by a machine unlearning algorithm from the loss value of the $x_{unseen}$.
To evaluate the forgetting performance, we assume a potential attacker.
The goal of the attacker is to obtain a piece of information about whether specific data $x$ was used in the training phase (belonging to the $\mathcal{D}_{train}$).
Thus, ideally, the binary classification model $\psi(\cdot)$ performs as follows:
\vspace{-0.1cm}
\begin{align}
    \psi(x) =
    \begin{cases}
        1 & \text{if } x \in \mathcal{D}_{forget} \\
        0 & \text{if } x \in \mathcal{D}_{unseen}
    \end{cases}
\end{align}

If the accuracy of $\psi(\cdot)$ is 0.5, the machine unlearning algorithm perfectly performs, which indicates that the $x_{forget}$ samples are not distinguishable from the $x_{unseen}$ samples.
We define $M$ as accuracy of $\psi(\cdot)$.
Finally, we also define the \textbf{forgetting score} as $(M-0.5)$ where the lower is better.

\subsubsection{Normalized Machine Unlearning Score}

The goal of machine unlearning algorithms is to obtain high scores for the aforementioned two metrics (1) model utility and (2) forgetting score simultaneously.
Thus, we introduce a new metric, Normalized Machine Unlearning Score (\textbf{NoMUS}) that is a comprehensive machine unlearning performance metric:
\vspace{-0.1cm}
\begin{align}
    P(\hat{y} = y)\times\lambda + (1 - abs(M - 0.5)\times2)\times(1-\lambda)
\end{align}

For $M$, the closer to 0.5 is the better.
Moreover, the $abs(\cdot)$ denotes the absolute value function.
We note that the $\lambda$ should be between 0 and 1 as a real number scalar.
If the $\lambda$ is 0.5, we samely weight (1) model utility and the (2) forgetting score.
The best score of NoMUS is 1 and the lowest score is 0 since the NoMUS is normalized to the real number scalar, constrained to $[0, 1]$.

\begin{figure}[htp]
\centering
      {\includegraphics[width=6.5cm]{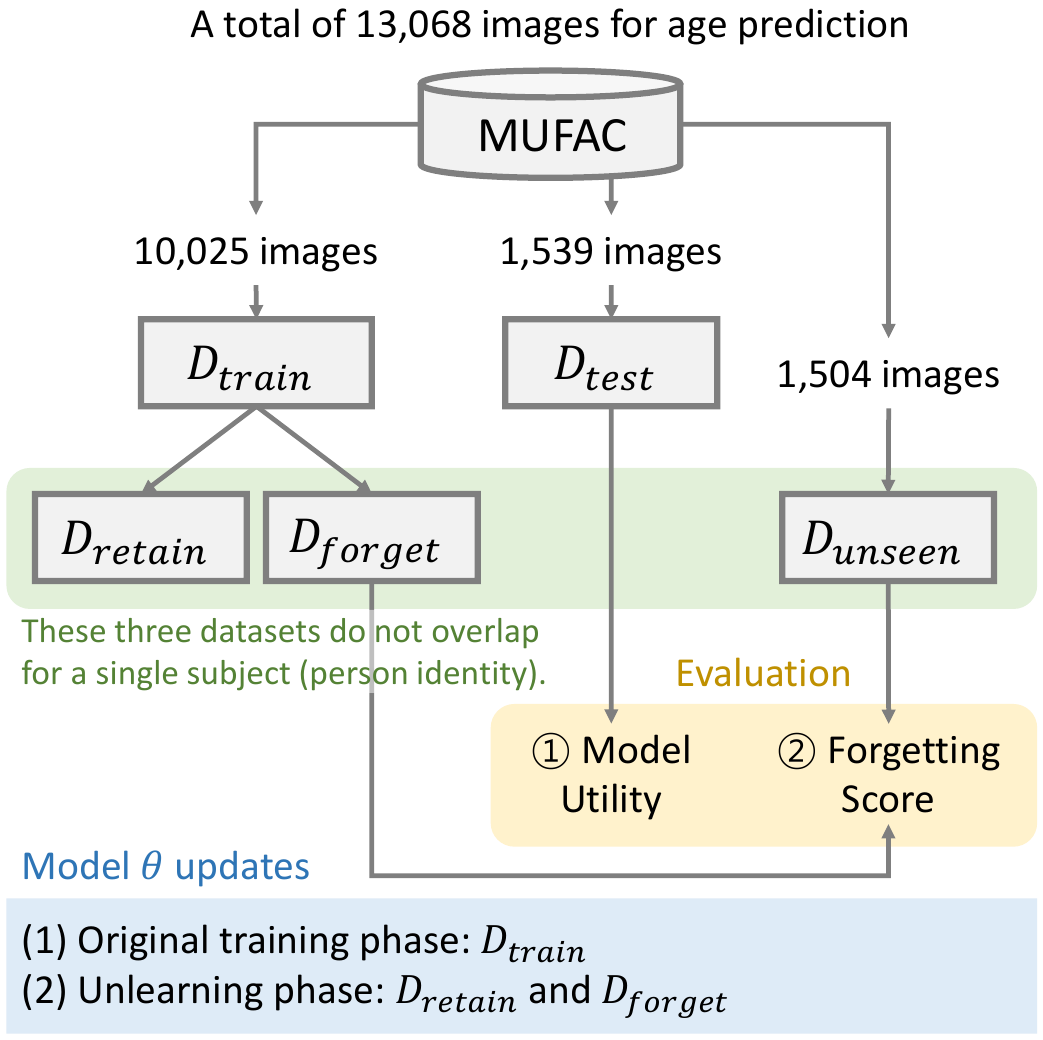}}
    \caption{The illustration of our MUFAC benchmark.}
    \label{fig:machine_unlearning_concept_2}
\end{figure}
\vspace{-0.1cm}

\begin{table*}[ht]
\centering
\caption{Overall performance of various machine unlearning algorithms on our \textit{MUFAC} dataset with $\lambda=1/2$. We emphasize the best score using \textbf{boldface} and the second best score using \textit{italics}. In the forgetting score, the lower is better.}
\vspace{-0.1cm}
\renewcommand{\arraystretch}{1.0}
\begin{adjustbox}{width=17.0cm, center}
\begin{tabular}{c |c|c| c|c|c|c|c|c|c|c}
\hline
&  \multirow{2}{*}{Evaluation Metrics} &  \multirow{2}{*}{Original} &  \multirow{2}{*}{Retrained} &  \multirow{2}{*}{Fine-tuning} &  \multirow{2}{*}{CF-3} &  \multirow{2}{*}{NegGrad} & \multicolumn{2}{c|}{UNSIR} &  \multirow{2}{*}{SCRUB} & \multirow{2}{*}{\shortstack{\textbf{Advanced}\\\textbf{NegGrad}}} \\
\cline{8-9}
& & & & &  & & Stage 1 & Stage 2 & & \\
\hline
\multirow{4}{*}{ResNet18}& Test Acc. &0.5951	&0.4880& \textbf{0.6055} &	0.5900&0.4048&0.5893	&0.5925	& \textit{0.5984}&0.5633 \\
& Top-2 Test Acc. & \textit{0.8804} &0.7667	&\textbf{0.8869}&	\textit{0.8804}&0.5932	&0.8778&	0.8674	&0.8745&0.8557\\
&Forgetting Score&0.2136	&\textbf{0.0445}	&0.2129	&0.2126	&\textit{0.0485}	 &0.2089&0.1990&0.1415&0.0953\\
\cline{2-11}
& Total Score (\textbf{NoMUS})&0.5839& \textbf{0.6994} &0.5898	&0.5823	&0.6538	&0.5857	&0.5972	& 0.6576 & \textit{0.6863}\\
\hline
\end{tabular}
\end{adjustbox} 
\label{tab:your_label}
\end{table*}
\vspace{-0.2cm}

\begin{table*}[ht]
\centering
\caption{Overall performance of various machine unlearning algorithms on our \textit{MUCAC} dataset with $\lambda=1/2$. We emphasize the best score using \textbf{boldface} and the second best score using \textit{italics}. In the forgetting score, the lower is better.}
\vspace{-0.2cm}
\renewcommand{\arraystretch}{1.0}
\begin{adjustbox}{width=17.0cm, center}
\begin{tabular}{c|c|c|c|c|c|c|c|c|c|c}
\hline
&  \multirow{2}{*}{Evaluation Metrics} &  \multirow{2}{*}{Original} &  \multirow{2}{*}{Retrained} &  \multirow{2}{*}{Fine-tuning} &  \multirow{2}{*}{CF-3} &  \multirow{2}{*}{NegGrad} & \multicolumn{2}{c|}{UNSIR} &  \multirow{2}{*}{SCRUB} & \multirow{2}{*}{\shortstack{\textbf{Advanced}\\\textbf{NegGrad}}} \\
\cline{8-9}
& & & & &  & & Stage 1 & Stage 2 & & \\
\hline
\multirow{3}{*}{ResNet18}& Average Test Acc.&0.8852	&0.8135	&0.9147	&\textit{0.9197}	&0.4193	&0.7087	& \textbf{0.9220} &0.9073	&0.7607\\
&Forgetting Score & 0.0568 & 0.0436	& 0.0708	& 0.0685	&0.0356	& \textit{0.0324} &0.0705	&0.0478 &\textbf{0.0152}\\
\cline{2-11}
& Total Score (\textbf{NoMUS})&0.8858	&0.8631	&0.8865	& \textit{0.8913}	&0.6740 &0.8219	&0.8905	&\textbf{0.9058} &0.8651\\
\hline
\end{tabular}
\end{adjustbox} 
\label{tab:your_label}
\end{table*}
\vspace{-0.2cm}

\subsection{Benchmark Datasets}

\subsubsection{Machine Unlearning for Facial Age Classifier}
We first present the \textit{MUFAC} dataset.
In this dataset, the original classification model solves the multi-class classification problem, the age estimation.
The age annotation consists of a total of 8 classes, from class 0 to class 7.
Each label represents a range of ages, for example, the \textit{class 0} covers the range between 0 years old to 6 years old.
We first divide the original dataset that contains a total of 11,564 images into $\mathcal{D}_{train}$ and $\mathcal{D}_{test}$.
Furthermore, we additionally construct the $\mathcal{D}_{unseen}$ that contains 1,504 images.
In the constructed datasets, we set any person (subject) to do not simultaneously belonging to the $\mathcal{D}_{train}$, $\mathcal{D}_{test}$ and $\mathcal{D}_{unseen}$.
After training the original model on $\mathcal{D}_{train}$, we finally split the $\mathcal{D}_{train}$ into the $\mathcal{D}_{forget}$ and $\mathcal{D}_{retain}$.
By doing so, anyone (personal identity) does not overlap across the $\mathcal{D}_{forget}$, $\mathcal{D}_{retain}$ and $\mathcal{D}_{unseen}$.
Thus, for the evaluation phase for machine unlearning, $\mathcal{D}_{forget}$ and $\mathcal{D}_{unseen}$ are successfully divided based on personal identity.
We note that the $\mathcal{D}_{test}$ is used for only the purpose of evaluating the (1) model utility and $\mathcal{D}_{unseen}$ is used for only the purpose of evaluating the (2) forgetting score in our work.

\subsubsection{Machine Unlearning for Celebrity Attribute Classifier}
We also introduce a new machine unlearning benchmark dataset, \textit{MUCAC}, by adjusting the CelebA dataset.
In the \textit{MUCAC} benchmark, the original model solves multi-label classification problems using binary labels (male/female, old/young, etc.).
As same as the \textit{MUFAC}, the \textit{MUCAC} benchmark also consists of $\mathcal{D}_{train}=\mathcal{D}_{forget}\cup\mathcal{D}_{retain}$, $\mathcal{D}_{test}$, and $\mathcal{D}_{unseen}$.
In the machine unlearning setup, these three dataset does not share the same personal identity (subject) utilizing the same methodology of the \textit{MUFAC}.

\section{Experiments}

To validate the usefulness of our two proposed benchmark datasets and evaluate the previously presented machine unlearning methods, we have experimented with these two datasets.
Firstly, the original models solve classification tasks according to their own purpose.
For preparing the original model $\theta_{original}$ for each dataset, we train various CNN models on the $\mathcal{D}_{train}$ of the two benchmarks respectively.
For example, we have observed that the standard ResNet18~\cite{ResNet} model can achieve a top-1 accuracy of 59.51\% for the \textit{MUFAC}.
For MUCAC, we have trained the multi-label classification models utilizing the three binary labels i.e., male/female, old/young, and smiling/unsmiling.
The classification models can simultaneously solve these three different types of binary classification tasks.
A multi-task ResNet18 model shows an average accuracy of 88.52\% for the three binary classification tasks.

After training the original model $\theta_{original}$ on the $\mathcal{D}_{train}$, we then evaluate the machine unlearning algorithms.
For this purpose, we divide the $\mathcal{D}_{train}$ into the $\mathcal{D}_{forget}$ and $\mathcal{D}_{retain}$.
In the machine unlearning phase, the machine unlearning algorithm can use $\mathcal{D}_{forget}$, $\mathcal{D}_{retain}$ and $\theta_{original}$ \textbf{as inputs of the algorithm}.
Basically, the goal of the machine unlearning algorithms is simultaneously (1) to memorize the retain dataset $\mathcal{D}_{retain}$ and (2) to unlearn the forget dataset $\mathcal{D}_{forget}$.
After the machine unlearning process, we finally (1) test the unlearned model $\theta_{unlearned}$ on the $\theta_{test}$ and also (2) train an additional classifier $\psi(\cdot)$ to evaluate the forgetting score on the $\theta_{forget}$ and $\theta_{unseen}$.
If the $\psi(\cdot)$ distinguishes the $\mathcal{D}_{retain}$ and $\mathcal{D}_{forget}$ well, the forgetting score is calculated as lower, indicating that some loss values of $x_{forget}$ are noticeable compared to the $x_{unseen}$.
We note that a good machine unlearning algorithm simultaneously shows high performance in the (1) model utility and (2) forgetting score.
Finally, we calculate the comprehensive score using our proposed, \textbf{NoMUS}, for each machine unlearning algorithm.

For experiments, we use various machine unlearning methods including Fine-tuning, NegGrad, CF, UNSIR, and SCRUB.
Fine-tuning adopts simply fine-tuning the $\theta_{original}$ on only the $\mathcal{D}_{retain}$ while the NegGrad indicates fine-tuning only on $\mathcal{D}_{forget}$ utilizing gradient ascent along the direction of increasing loss~\cite{FisherForgetting}.
Interestingly, we have found that our simple adjustment to the NegGrad (\textbf{Advanced NegGrad}) shows competitive performance in our benchmark settings compared to the previously state-of-the-art methods.
Our \textit{Advanced NegGrad} is a simple method that utilizes the joint loss of Fine-tuning and NegGrad in the same training batches.
We note that this simple adaptation frequently outperforms the recent SOTA methods in our new benchmark settings.
The detailed implementations are described in the supplementary materials.

\section{Conclusion}

In this paper, we introduce two new machine unlearning benchmark datasets, \textit{MUFAC} and \textit{MUCAC}.
Our presented datasets can be used for imitating real-world scenarios, which is different from the class-unlearning setup that has been broadly adopted in previous studies.
Moreover, we also provide all the machine unlearning performance of the state-of-the-art unlearning methods with extensive experiments.
Furthermore, we point out that some SOTA methods do not show modest performance in our proposed benchmark settings.
We believe that these results might be introduced due to fitting on the previously presented class-unlearning benchmark datasets.
We distribute all the datasets, codes, and trained models, expecting that our work will be helpful for the growth of the machine-learning research fields.

\section{Acknowledgements}
This research was supported by Brian Impact, a non-profit organization dedicated to advancing science and technology

\bibliography{aaai24}

\newpage
\onecolumn
\appendix

\section{Supplementary Material}

\subsection{Evaluation Details}

\subsubsection{Machine Unlearning Algorithm Evaluation Pipeline}

Various recent machine unlearning studies adopt the following steps~\cite{CatastrophicallyForgetting,ImageRetrieval,SCRUB,FederatedPruning,AlgorithmicStability,UnrollingSGD,Remember,Descent,Randomized,Adaptive}.
Firstly, we train an original model $\theta_{original}$ on the $\mathcal{D}_{train}$.
Then, we apply the machine unlearning algorithms to the trained original models $\theta_{original}$.
We note that, for a fair comparison, the $\theta_{original}$ should be equally used for all the machine unlearning algorithms in the machine unlearning evaluation pipeline.
Finally, we calculate the (1) model utility and (2) forgetting score over the unlearned models $\theta_{unlearned}$ to evaluate various machine unlearning methods.

\subsubsection{Task-Agnostic Machine Unlearning}

In this work, we identify the potential issue of the class-unlearning setting and pose a new machine-unlearning benchmark where the original task is not changed intrinsically after machine unlearning.
The class-unlearning setup removes specific classes completely as a result of the machine unlearning.
Instead, our task-agnostic unlearning setup pursues to maintain the original task and achieve good accuracy (classification performance) for the original task even after the unlearning is performed.
The proposed task-agnostic machine unlearning setup can be useful to evaluate the robustness of machine unlearning algorithms in various aspects.
Moreover, we claim that the task-agnostic machine unlearning setup tends to be well generalized to the real-world machine unlearning scenario.

\subsubsection{Details for Membership Inference Attack (MIA) and NoMUS}

Similarly to the previous studies, we underline the two most important metrics.
Calculating the forgetting score properly for a machine unlearning algorithm is crucial.
In our main experiment sections, we adopt the MIA as an evaluation metric while using the \textit{accuracy} for the model utility.
In our benchmarks, the $x_{forget}$ belongs to the $\mathcal{D}_{train}$.
The $\psi(\cdot)$ can be easily trained using the loss values of the input data $x$ and the corresponding labels $y$.
In this procedure, for training an additional binary classifier $\psi(\cdot)$ (MIA attacker), the data pair ($x$, $y$) should be sampled from the $\mathcal{D}_{forget} \cup \mathcal{D}_{unseen}$ for properly evaluating the forgetting score.
In conclusion, the accuracy of this binary classification model $\psi(\cdot)$ represents the attack success rate of the MIA.
For $M$, the closer to 0.5 is the better and the $M$ denotes the accuracy of $\psi(\cdot)$.
We note that the $\lambda$ should be between 0 and 1 as a real number scalar.
As aforementioned, when the $\lambda$ is 0.5, we samely weight (1) model utility and the (2) forgetting score.
If the $\lambda$ is 1, we consider only the model utility.
In contrast, we consider only the forgetting score if the $\lambda$ is 0.
These normalized metrics that combine the two different measurements simultaneously can be useful to evaluate the machine-unlearning performance.
The best score of our proposed NoMUS is 1 and the lowest score is 0 since the NoMUS is normalized to the real number scalar, constrained to $[0, 1]$.

\subsection{Adjustment of the Existing Methods}

For experiments, we adopt the most representative machine unlearning methods including CF-k~\cite{CatastrophicallyForgetting}, UNSIR~\cite{UNSIR}, and SCRUB~\cite{SCRUB}.
However, the recently proposed machine unlearning methods generally adopt the class-unlearning setting~\cite{CatastrophicallyForgetting,UNSIR,SCRUB}.
Thus, we adjust some methods to calculate the performance for the task-agnostic setting.
If an original machine unlearning method aims to generate synthesized data that maximizes the loss of a specific \textit{classes to forget}, we can instead synthesize data that maximizes the distance between the synthesized data and the \textit{instances to forget} as an advanced adjustment for that methods.
For example, to implement the UNSIR, we utilize the following formulation:

\begin{align}
    {\argmax}_{x_{gen}} d(F(x_{gen}), F(x_{forget})) + r(x_{gen})
\end{align}

where the $x_{gen}$ denotes the generated data by a machine unlearning algorithm.
For UNSIR~\cite{UNSIR}, the $x_{gen}$ could be a noise image.
Moreover, the $r(\cdot)$ indicates a regularization function that regularizes the $x_{gen}$ to be constrained to a certain pixel value bound.
For noise generation like UNSIR~\cite{UNSIR}, this regularization term can be useful to make the magnitude of the pixels of the noise data not increase too much.
The $d$ denotes the distance measure function.
For calculating the semantic distance between the two images, we extract the logit feature vectors by forwarding the original image $x$ into the feature extractor parts $F(\cdot)$ (before softmax) of the trained model $\theta$.
Because UNSIR~\cite{UNSIR} does not adopt any additional network, we utilize the $\theta_{unlearned}$ itself and extract the logit vectors by forwarding the inputs into the model.
Then, we can calculate the $l_2$ distance between these two images in the feature space.

\subsection{Performance of Single-Task Models and Visualization for Datasets}

\begin{figure*}[htp]
    \centering   \centerline{\includegraphics[width=0.8\textwidth]{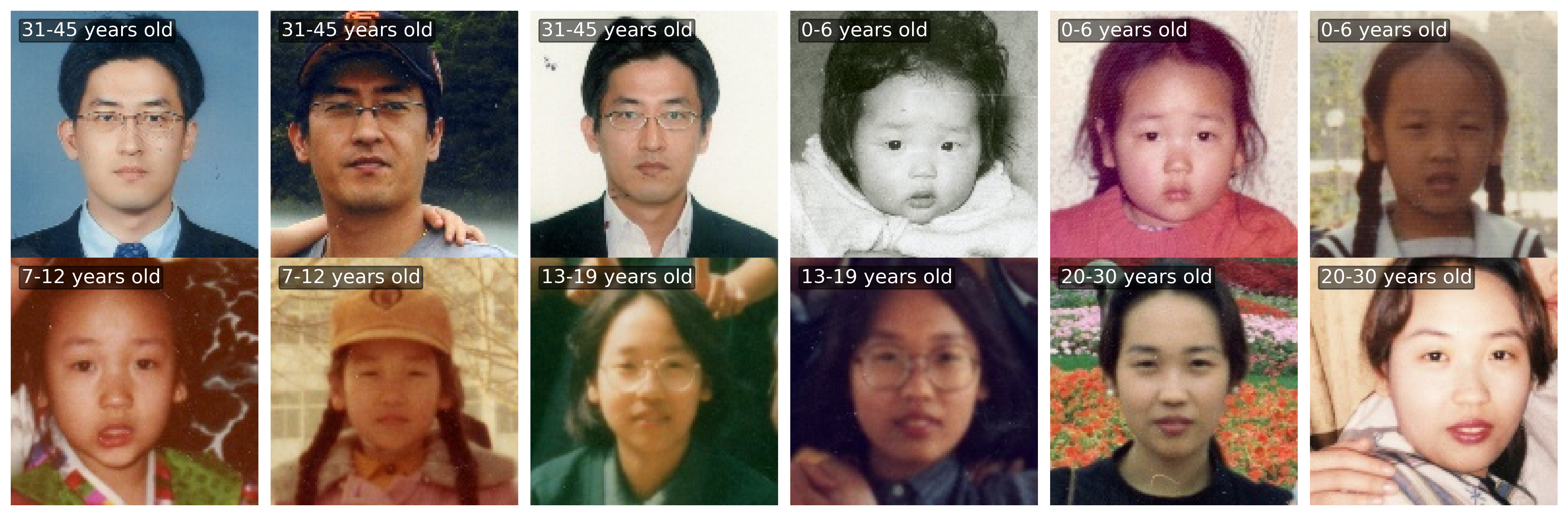}}
    \caption{Example images that are sampled from our presented \textit{MUFAC} dataset.}
    \label{fig:MUFAC}
\end{figure*}

\begin{figure*}[htp]
    \centering   \centerline{\includegraphics[width=0.8\textwidth]{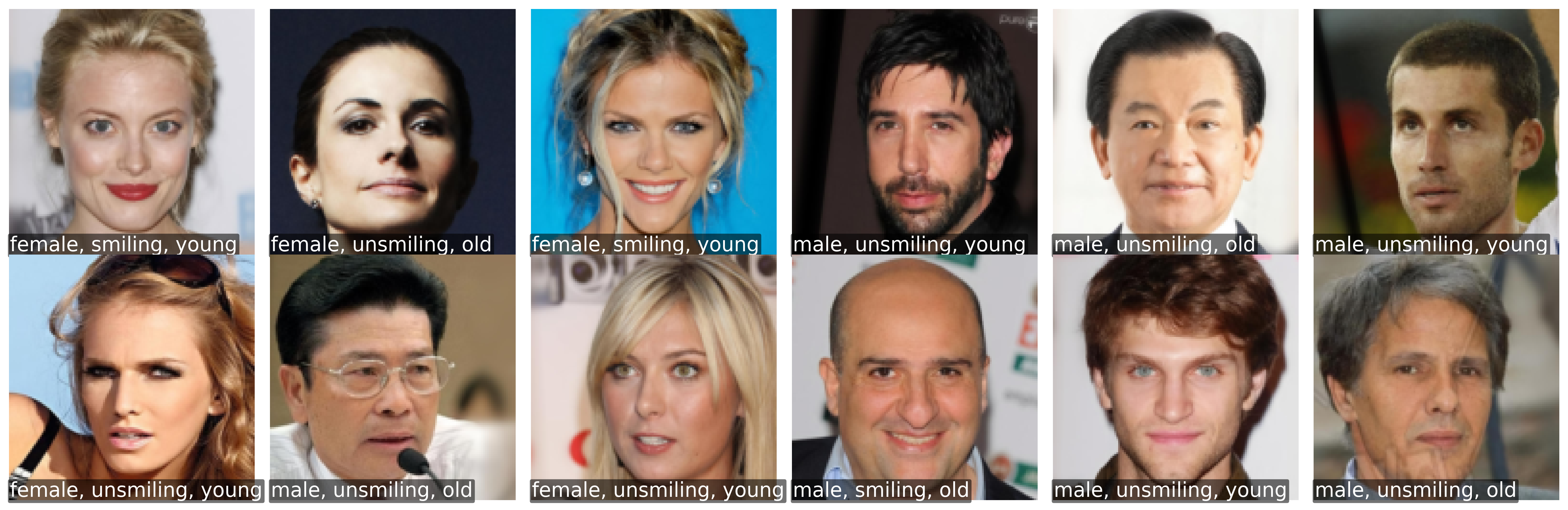}}
    \caption{Example images that are sampled from our presented \textit{MUCAC} dataset.}
    \label{fig:MUCAC}
\end{figure*}

We also report the performance of single-task models on the proposed \textit{MUCAC} dataset.
The \textit{MUCAC} provides three types of binary labels, i.e., male/female, old/young, and smiling/unsmiling respectively for each classification task.
Therefore, we have also trained the binary classification models that address individually each facial recognition task.
Because we present two datasets, \textit{MUFAC} and \textit{MUCAC}, we illustrate some sample images from these two datasets for the convenience of the readers.
As shown in the Figure~\ref{fig:MUFAC}, the \textit{MUFAC} dataset contains data pairs ($x$, $y=age$).
Moreover, the \textit{MUCAC} dataset consists of data ($x$, $y^{1}=gender$, $y^{2}=smiling$, $y^{3}=age$) similarly to the \textit{MUFAC} as shown in Figure~\ref{fig:MUCAC}.
Thus, we can use the \textit{MUCAC} benchmark for evaluating machine unlearning methods, when the original model solves the binary classification problems.

\begin{table*}[ht]
\centering
\caption{Overall performance of various machine unlearning algorithms on our \textit{MUCAC} male/female classification (binary classification) benchmark with $\lambda=1/2$. We emphasize the best score using \textbf{boldface} and the second best score using \textit{italics}.}
\vspace{-0.2cm}
\renewcommand{\arraystretch}{1.3}
\begin{adjustbox}{width=17.0cm, center}
\begin{tabular}{c |c|c| c|c|c|c|c|c|c|c}
\hline
&  \multirow{2}{*}{Evaluation Metrics} &  \multirow{2}{*}{Original} &  \multirow{2}{*}{Retrained} &  \multirow{2}{*}{Fine-tuning} &  \multirow{2}{*}{CF-3} &  \multirow{2}{*}{NegGrad} & \multicolumn{2}{c|}{UNSIR} &  \multirow{2}{*}{SCRUB} & \multirow{2}{*}{\shortstack{\textbf{Advanced}\\\textbf{NegGrad}}} \\
\cline{8-9}
& & & & &  & & Stage 1 & Stage 2 & & \\
\hline
\multirow{3}{*}{ResNet18}& Test Acc. &0.9835	&0.9515	&\textbf{0.9849}	&0.9840	&0.1762	&0.9481	&\textit{0.9845}	&0.1762	&0.9147\\
&Forgetting Score&0.0306&\textit{0.0154}&0.0281	&0.0291	&0.1289	&0.0638	&0.0481	&0.1329 &\textbf{0.0129}\\
\cline{2-11}
& Total Score&0.9611&0.9603	&\textbf{0.9643}&\textit{0.9629}&	0.4592 &0.9102&	0.9441	&0.4552 & 0.9444\\
\hline
\end{tabular}
\end{adjustbox} 
\label{tab:your_label}
\end{table*}

\begin{table*}[ht]
\centering
\caption{Overall performance of various machine unlearning algorithms on our \textit{MUCAC} smiling/unsmiling classification (binary classification) benchmark with $\lambda=1/2$. We emphasize the best score using \textbf{boldface} and the second best score using \textit{italics}.}
\vspace{-0.2cm}
\renewcommand{\arraystretch}{1.3}
\begin{adjustbox}{width=17.0cm, center}
\begin{tabular}{c |c|c| c|c|c|c|c|c|c|c}
\hline
&  \multirow{2}{*}{Evaluation Metrics} &  \multirow{2}{*}{Original} &  \multirow{2}{*}{Retrained} &  \multirow{2}{*}{Fine-tuning} &  \multirow{2}{*}{CF-3} &  \multirow{2}{*}{NegGrad} & \multicolumn{2}{c|}{UNSIR} &  \multirow{2}{*}{SCRUB} & \multirow{2}{*}{\shortstack{\textbf{Advanced}\\\textbf{NegGrad}}} \\
\cline{8-9}
& & & & &  & & Stage 1 & Stage 2 & &  \\
\hline
\multirow{3}{*}{ResNet18}& Test Acc. &0.9467&0.6518	&\textit{0.9476}	&0.9472	&0.5549 &0.8619	&\textbf{0.9506}&0.5549 &0.9423\\
&Forgetting Score&0.0346&0.0182	&\textit{0.0279}	&0.0294	&0.0366	&0.0416	&\textbf{0.0271}&0.4680 &0.0354\\
\cline{2-11}
& Total Score&0.9387&0.8077	&\textit{0.9459}	&0.9442	&0.7408	&0.8893	&\textbf{0.9482}	&0.3094 &0.9357\\
\hline
\end{tabular}
\end{adjustbox} 
\label{tab:your_label}
\end{table*}

\begin{table*}[ht]
\centering
\caption{Overall performance of various machine unlearning algorithms on our \textit{MUCAC} young/old classification (binary classification) benchmark with $\lambda=1/2$. We emphasize the best score using \textbf{boldface} and the second best score using \textit{italics}.}
\vspace{-0.2cm}
\renewcommand{\arraystretch}{1.3}
\begin{adjustbox}{width=17.0cm, center}
\begin{tabular}{c |c|c| c|c|c|c|c|c|c|c}
\hline
&  \multirow{2}{*}{Evaluation Metrics} &  \multirow{2}{*}{Original} &  \multirow{2}{*}{Retrained} &  \multirow{2}{*}{Fine-tuning} &  \multirow{2}{*}{CF-3} &  \multirow{2}{*}{NegGrad} & \multicolumn{2}{c|}{UNSIR} &  \multirow{2}{*}{SCRUB} & \multirow{2}{*}{\shortstack{\textbf{Advanced}\\\textbf{NegGrad}}} \\
\cline{8-9}
& & & & &  & & Stage 1 & Stage 2 & & \\
\hline
\multirow{3}{*}{ResNet18}&Test Acc. &0.9089&0.8271	&\textbf{0.9147}	&\textit{0.9118}	&0.1733	&0.8260	&0.9021	&0.8929 &0.5573 \\
&Forgetting Score&0.0456&0.0234	&0.0426	&0.0456	&0.0513	&\textit{0.0229}	&0.0493	&0.0428 &\textbf{0.0139}	\\
\cline{2-11}
& Total Score&0.9088&0.8901	&\textbf{0.9147}	&\textit{0.9103}&0.5353 &0.8901	&0.9017	&0.9036 &0.7647\\
\hline
\end{tabular}
\end{adjustbox} 
\label{tab:your_label}
\end{table*}

\vspace{-0.5cm}

\end{document}